\documentclass[lettersize,journal]{IEEEtran}
\usepackage{amsmath,amsfonts}
\usepackage{algorithmic}
\usepackage{algorithm}
\usepackage{array}
\usepackage[caption=false,font=normalsize,labelfont=sf,textfont=sf]{subfig}
\usepackage{textcomp}
\usepackage{stfloats}
\usepackage{url}
\usepackage{verbatim}
\usepackage{graphicx}
\usepackage{cite}
\usepackage{pifont} 
\usepackage{bm}
\usepackage{booktabs}

\newcommand{\vecv}{v}
\newcommand{\vect}{\tau}
\newcommand{\matt}{\bm{\tau}}

\begin{document}

\title{FAAGC: Feature Augmentation on Adaptive Geodesic
    Curves based on the shape space theory}

\author{
    Yuexing Han*, and Ruijie Li
    \thanks{*Corresponding author: Yuexing Han (e-mail: han\_yx@i.shu.edu.cn).}
    \thanks{Yuexing Han is with the School of Computer Engineering and Science, Shanghai University, 99 Shangda Road, Shanghai 200444, People's Republic of China,
    and also with the Key Laboratory of Silicate Cultural Relics Conservation (Shanghai University), Ministry of Education (e-mail: han\_yx@i.shu.edu.cn).}
    \thanks{Ruijie Li is with the School of Computer Engineering and Science, Shanghai University, 99 Shangda Road, Shanghai 200444, People's Republic of China (e-mail: liruijieas@163.com).}
}




\maketitle

\begin{abstract}
    Deep learning models have been extensively adopted in various application domains. However, their reliance on massive amounts of training data restricts feasibility in fields where only limited samples can be obtained. Data augmentation is one of the key techniques to mitigate these issues by artificially expanding training samples while preserving data distribution, thereby improving model generalization. Nevertheless, most conventional augmentation methods are modality-specific and often fail to fully utilize the non-linear properties of features, particularly under extreme data scarcity.
    To overcome these limitations, we propose a feature augmentation module named FAAGC (Feature Augmentation on Adaptive Geodesic Curves) to augment training data at the feature level. In the FAAGC module, features are extracted by a deep neural network into the pre-shape space. With the shape space theory, we adaptively construct a Geodesic curve to closely fit the distribution of feature vectors in the pre-shape space for each class.
    Feature augmentation is then performed by sampling along the Geodesic curve. Extensive experiments demonstrate that FAAGC improves image classification accuracy under data-scarce conditions and generalizes well to various benchmark datasets.
\end{abstract}


\section{Introduction}
Using deep neural networks to extract and transform features has become a mainstream approach for various data modalities and downstream tasks. The powerful representational capacity of deep neural networks for complex patterns emerges particularly when sufficient training data is available.
However, data scarcity remains a significant challenge, especially in fields where acquiring large volumes of high-quality labeled data can be expensive and time-consuming~\cite{litjens2017survey,butler2018machine}. If trained on insufficient samples, deep neural networks frequently suffer from overfitting, resulting in degraded performance on test datasets.
To alleviate these problems, data augmentation~\cite{krizhevsky2012imagenet,shorten2019survey} has been extensively adopted in deep learning pipelines. Data augmentation aims to enhance model generalization by artificially increasing training samples while maintaining the original data distribution. For image processing, label-preserving transformations such as random cropping and image flipping~\cite{krizhevsky2012imagenet} are frequently employed as standard augmentation practices.

While data augmentation strategies are beneficial, they have been specifically designed for particular datasets and tasks, often requiring guidance from domain experts. The variability of strategies limits the model's generalizability during training and preprocessing.
For example, augmentation techniques such as random rotations or flips may improve performance on natural image datasets like CIFAR-10, but can be detrimental or even introduce artifacts when applied to medical imaging tasks, where orientation and spatial relationships are critical~\cite{shorten2019survey}.
Furthermore, when training samples are extremely scarce, common input transformations produce features with limited complexity, resulting in the model failing to capture the complete data distribution.

Feature-level data augmentation provides a more general alternative to augment raw input data. For instance, DeVries and Taylor~\cite{devries2017dataset} proposed a feature-level data augmentation approach based on the feature space and discovered a higher probability of encountering real samples in the feature space than in input space when traversing along a manifold.
Most feature-level data augmentation methods in deep learning, such as Manifold Mixup~\cite{verma2019manifold} and Feature-level SMOTE~\cite{liu2024feature}, have been proposed to generate synthetic features by linear interpolation, feature mixing, or distribution sampling. However, such practices may neglect the complex structure present in feature spaces. As a result, they may fail to preserve intrinsic semantic consistency or exploit the full potential of feature distributions, especially under severe data scarcity.

To address these issues, we propose a Feature Augmentation on Adaptive Geodesic Curves (FAAGC) feature augmentation module, based on the shape space theory~\cite{kendall1984shape,han2010recognition}. According to the shape space theory, objects sharing the same shape are represented as vectors lying on a great circle in the pre-shape space. "Adaptive" means that for each class, the Geodesic curve is optimized according to the distribution of sample features, rather than relying on fixed or randomly selected endpoints.

Assuming that features extracted by deep learning models capture critical information, we project these features into the pre-shape space and adaptively construct a segment of great circle for each class, enabling Geodesic-based interpolation that preserves image information. By sampling features along the Geodesic curve, we generate augmented features to enhance classifier performance.

The approach of augmenting features in the pre-shape space along the Geodesic curve, known as FAGC, was first proposed in~\cite{han2023gcfa}. However, FAGC employs an iterative algorithm that is time-consuming and does not yield optimal performance, thus limiting its practical applicability. FAAGC overcomes these limitations by carefully redesigning both the loss function and computational pipeline, and by leveraging gradient descent to optimize the Geodesic Curve Fitting efficiently, thereby markedly improving computational efficiency and overall performance.

FAAGC effectively augments scarce data while maintaining applicability across diverse data modalities. Experimental results show that FAAGC consistently improves performance on visual datasets—both with and without image augmentation—and provides additional performance gains when integrated with other methods of image augmentation.

Our key contributions are summarized as follows:
\begin{itemize}
    \item We propose a FAAGC module that augments samples along the Geodesic curve in the pre-shape space. The generated feature vectors match the original data distribution more closely.
    \item We optimize the computation of the loss function designed for the FAAGC module, significantly reducing training time and achieving superior augmentation performance compared to prior methods.
    \item Extensive experiments and ablation studies on multiple vision benchmark datasets demonstrate that the FAAGC module effectively augments features under data-scarce conditions, while providing additive gains when combined with the conventional image augmentation approaches.
\end{itemize}

\section{Related Works}

\subsection{Image Data Augmentation}

To improve model performance and generalization under limited sample sizes, researchers have proposed multiple data augmentation methods on image domain.

In computer vision, commonly used data augmentation techniques include flipping, rotation, translation, scaling, noise addition, occlusion, and color jittering. These operations are typically applied to input images either individually or in sequence, according to predefined settings of probabilities, orders, and magnitudes. Such transformations increase the diversity of training samples and improve generalization ability~\cite{shorten2019survey}.

Despite success, data augmentation methods above require expertise and manual work to design policies that capture prior knowledge in each domain.
AutoAugment~\cite{cubuk2019autoaugment} is proposed to formulate data-augmentation policy search as a discrete optimization problem and uses a reinforcement learning controller to find transformation chains that maximize validation accuracy. RandAugment~\cite{cubuk2020randaugment} is a simplified approach that randomly selects augmentation operations with a uniform magnitude, enabling more flexible and efficient policy discovery. AugMix~\cite{hendrycks2019augmix} is proposed to enhance diversity by mixing multiple stochastically sampled augmentation chains, improving both robustness and generalization.

For high-precision tasks like medical imaging analysis, data augmentation involves generating synthetic samples using Generative Adversarial Networks (GANs), simulating realistic lesion characteristics, creating rare case images, and augmenting CT and MRI data with specific slice augmentations. These methods significantly enhance data diversity and model robustness~\cite{frid2018synthetic}. While GAN-based augmentation can be effective, it becomes challenging when sample sizes are very limited due to its reliance on abundant data for stable training. Insufficient samples often lead to issues like mode collapse or overfitting, hindering the generation of high-quality synthetic data~\cite{karras2020training}.
In specialized fields such as medicine, both general-purpose and domain-specific data augmentation methods can be utilized to enhance task performance~\cite{athalye2023domain}. However, these augmentation methods require validation through dataset-specific experiments and the acceptance of domain experts to ensure their applicability and effectiveness.

In contrast, the following methods perform data augmentation entirely at the feature level to improve the generalization of deep learning models.
Goodfellow et al.~\cite{goodfellow2014explaining} pioneered the use of adversarial examples, revealing model vulnerabilities and subsequently proposing feature space perturbations to improve model robustness.
DeVries and Taylor~\cite{devries2017dataset} leveraged Variational Autoencoders (VAEs)~\cite{kingma2013auto} to embed data in a latent space, with the application of extrapolation, interpolation, and perturbation to generate diverse augmented samples.
Building on the idea of feature mixing, Verma et al.~\cite{verma2019manifold} developed Manifold Mixup, which interpolates hidden representations to improve robustness and accuracy against adversarial attacks.
Simple yet effective strategies have also emerged. For instance, Li et al.~\cite{li2021simple} demonstrated that simply adding certain noise to sample features can substantially boost generalization and robustness in transfer learning scenarios. Li et al.~\cite{li2021feature} also introduced MoEx, a technique that creates new samples by exchanging the mean and variance of features between different examples.
Manifold-aware generative methods have been explored as well: Chadebec et al.~\cite{chadebec2021data} utilized manifold-constrained VAE sampling for more effective augmentation.
Other domain-specific advancements include feature-level augmentation for improving long-tailed data distributions~\cite{chu2020feature}, and the extension of SMOTE to the feature space by Liu et al.~\cite{liu2024feature} to enrich minority classes in fault diagnosis tasks.

\subsection{The Shape Space Theory}
The shape space theory, introduced by Kendall~\cite{kendall1984shape, kendall2009shape}, was originally proposed to describe objects' shapes and their equivalent transformations in non-Euclidean spaces, namely the pre-shape space and the shape space. In the pre-shape space, variations in position and scale of shapes are ignored, whereas in the shape space, variations in position, scale, and orientation are all disregarded.
The shape space theory is later applied to image processing, such as object recognition, where the distance between two objects in the shape space determines recognition outcomes~\cite{han2010recognition}.

To enable computational representation and analysis, a object \( \vecv \) is represented with a set of \( d \) landmarks in the Euclidean space, denoted as

\begin{equation}
    \vecv = \bigl[(x_1, y_1), \ldots, (x_i, y_i) \ldots, (x_d, y_d)\bigr],
\end{equation}
where \((x_i,y_i)\) denotes the coordinate of the \(i\)-th landmark of the object.
To exclude the effect of position, we subtract the mean value of each coordinate dimension \(x\) and \(y\), getting the result vector \(\vecv'\):
\begin{equation}
    \label{eq:sub_mean}
    \vecv^{\prime} =
    \bigl[(x_1- \bar{x}, y_1- \bar{y}), \ldots, (x_i- \bar{x}, y_i- \bar{y}) \ldots, (x_d- \bar{x}, y_d- \bar{y})\bigr]
\end{equation}
where \( \bar{x} \) and \( \bar{y} \) are the means of \( \{x_i\} \) and \( \{y_i\} \), respectively.

Next, we exclude the scale and denote the resulting vector \(\vect \) as the pre-shape of \( \vecv \):

\begin{equation}
    \vect = \frac{\vecv'}{\| \vecv' \|} ,
    \label{eq:projection}
\end{equation}
where \( \| \cdot \| \) denotes the Euclidean norm.

The projected pre-shape vector $\vect$ lies on the unit hypersphere zero-mean coordinates along both the $x$ and $y$ axes, resulting in $2d-3$ degrees of freedom, i.e., $\vect \in S_*^{2d-3}$.
All changes of rotation, scaling, and position of the $\vecv$, form an orbit denoted by $O(\vect)$. The set of all shapes on the unit hyper-sphere forms the orbit space \(\Sigma_2^d\), i.e. a shape space, which is described as:

\begin{equation}
    \Sigma_{2}^{d} = \{ O(\vect): \vect \in S_{*}^{2d-3}\}.
\end{equation}

The set of all points on a great circle can be used to represent a specific shape. For any two vectors \(\vect_{1}, \vect_{2} \in S_{*}^{2d-3} \), their Geodesic distance is calculated as the great circle distance \(d(\vect_{1}, \vect_{2})\)~\cite{kendall2009shape}:

\begin{equation}
    \label{eq:preshape_distance}
    d(\vect_{1}, \vect_{2}) = \arccos{\langle \vect_{1}, \vect_{2} \rangle},
\end{equation}
where \(\langle \vect_{1}, \vect_{2} \rangle\) denotes the inner product of \(\vect_{1}\) and \(\vect_{2}\).

The distance  \( d_p \) measures the distance between two objects in the shape space and can be computed as the minimal Geodesic distance between the great circles associated with the projections of their features in the shape space \( \Sigma_{2}^{d} \), expressed as:

\begin{equation}
    \label{eq:dp}
    d_{p}\bigl[\,O(\vect_{1}),\,O(\vect_{2})\bigr] =
    \inf\Bigl\{ d(\alpha, \beta) \;:\;\alpha \in O(\vect_{1}),\;\beta \in O(\vect_{2}) \Bigr\}.
\end{equation}

By constructing a Geodesic curve between two vectors in the pre-shape space \(S_{*}^{2d-3}\), new data in the pre-shape space can be sampled along the curve. Given two feature vectors \(\vect_{\mathrm{start}} , \vect_{\mathrm{end}} \in S_{*}^{2d-3}\), intermediate vectors along the Geodesic curve can be generated according to~\cite{han2010recognition}:


\begin{multline}
    \Gamma_{(\vect_{\mathrm{start}}, \vect_{\mathrm{end}})}(s)
    = (\cos s) \cdot \vect_{\mathrm{start}} \label{eq:intermediatepoints} \\
    \quad+ (\sin s) \cdot
    \frac{\vect_{\mathrm{end}}
        - \vect_{\mathrm{start}} \cdot
        \cos\!\big(\theta_{(\vect_{\mathrm{start}}, \vect_{\mathrm{end}})}\big)}
    {\sin\!\big(\theta_{(\vect_{\mathrm{start}}, \vect_{\mathrm{end}})}\big)} , \\
    0 \le s \le \theta_{(\vect_{\mathrm{start}}, \vect_{\mathrm{end}})}
\end{multline}

where \(\theta_{(\vect_{\mathrm{start}}, \vect_{\mathrm{end}})} \) denotes the Geodesic distance between \(\vect_{\mathrm{start}}\) and \(\vect_{\mathrm{end}}\)
, and \(s\) represents the angle from \(\vect_{\mathrm{start}}\) to the sampled point \(\Gamma_{(\vect_{\mathrm{start}}, \vect_{\mathrm{end}})}(s) \) along the Geodesic curve.

The shape space theory can be used in the area of deep learning. Many studies start to use image features extracted by deep neural networks as representations of the shape-related features of key objects in corresponding object, and apply the shape space theory to enhance data.
For instance, Vadgama et al.~\cite{vadgama2022kendall} tweaked a VAE so that MNIST images are first mapped into latent where rotation is separated, allowing the same digit to be reconstructed with new angles.
The GLASS~\cite{muralikrishnan2022glass} method iteratively augments the 3D model dataset by alternating between training a VAE and exploring random perturbations in its low-dimensional latent space, guided by the ARAP geometric deformation energy, thereby generating a large amount of high-quality data.
While the above approaches requires a dedicated backbone to generate entirely new samples, Han et al.~\cite{han2023gcfa} proposed the FAGC framework, which leverages features extracted from pre-trained image models as landmarks to represent shapes, projects them as key points into the pre-shape space, and performs feature-level augmentation to improve classification accuracy in low-data scenarios.
However, with the FAGC framework, each iteration requires computing distances between all data points and every sampled point along the Geodesic curve. This process causes the runtime to increase dramatically as the sample size grows.

These approaches highlight the potential of the shape space theory in data augmentation, especially for low-resource tasks across various domains.

\section{The FAAGC module}
\label{sec:FAAGC}

To address the problem of insufficient sample size of the dataset and the risk of overfitting during model training, we introduce the FAAGC module to augment data at the feature level based on the shape space theory.

\begin{figure*}[!t]
    \centering
    \includegraphics[width=1\linewidth]{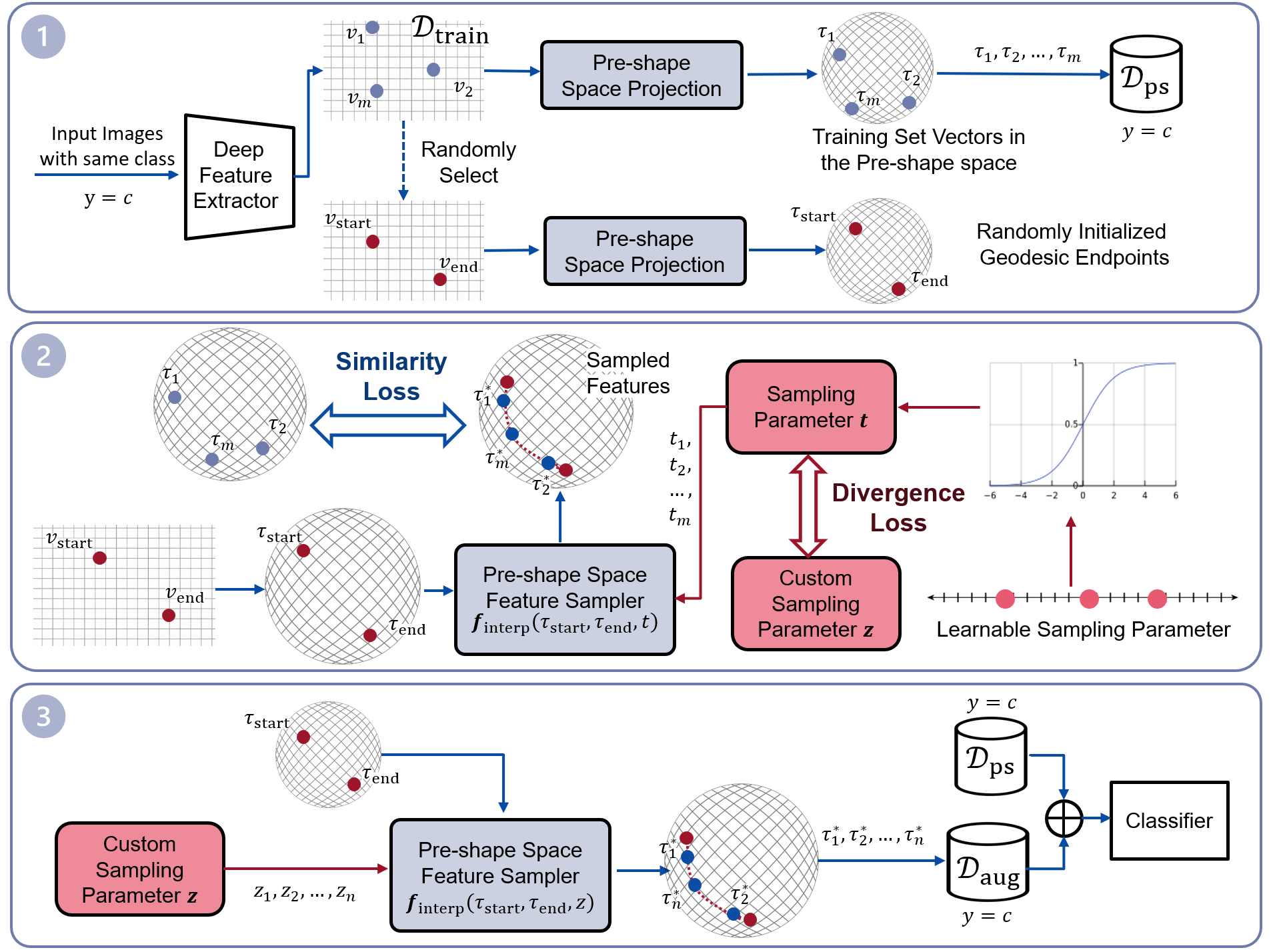}
    \caption{Workflow for Feature Augmentation on Adaptive Geodesic Curves (FAAGC)}
    \label{fig:adaptive-geodesic}
\end{figure*}

\subsection{Sample Projection into Pre-shape Space}
\label{sec:feat_extraction}

As illustrated in \ding{172} of the Fig.~\ref{fig:adaptive-geodesic}, the FAAGC algorithm initiates by projecting training samples into the pre-shape space.

For input images represented by tensors of size \(H \times W \times C\), where \(H\), \(W\), and \(C\) may vary across images, we first employ a vision model to embed each image into a feature vector in the Euclidean space.
Because the pre-trained neural network models are able to extract highly discriminative image features,  neural network models, such as the Vision Transformers~\cite{dosovitskiy2020vit}, are adopted as the feature extractor.
Thus, the image is projected from \(\mathbb{R}^{H \times W \times C}\) to \(\mathbb{R}^{d}\), where \(d\) denotes the dimensionality of the vector extracted from the image by the feature extractor.
The resulting \(d\)-dimensional vector is subsequently projected into the pre-shape space.

For labeled training data, we organize the extracted features and their corresponding labels into class-wise datasets:
\begin{equation}
    \label{eq:feature_dataset}
    \mathcal{D}_{\mathrm{train},c} = \left\{ (\vecv_{i}, c) \mid i = 1, \ldots, m \right\}, c = 1, \ldots, N_c,
\end{equation}
where $\vecv_{i} \in \mathbb{R}^d$ denotes the extracted feature vector of the $i$-th image sample with label $c$, $m$ denotes the number of samples per class, and $N_c$ is the total number of classes.
The following content always assumes the image samples are from the same label. Thus, $\mathcal{D}_{\mathrm{train},c}$ can be simplified as $\mathcal{D_{\mathrm{train}}}$ in the subsequent derivations.

According to the shape space theory, the landmarks of a shape structure typically required to be two-dimensional or three-dimensional.
Consequently, before projecting a \(d\)-dimensional feature vector, which is obtained with the feature extractor, into the pre-shape space, we first expand its dimensionality.
Using the up-dimension procedure proposed by Han et al.~\cite{han2023gcfa} we duplicate each feature dimension for the \(i\)-th feature vector:
\begin{equation}
    \label{eq:duplicate}
    \vecv_i' = \bigl[(v_i[1], v_i[1]),\ldots,(v_i[d],v_i[d])\bigr],
\end{equation}
where \(v_i[\cdot]\) denotes an element in the feature vector \(\vecv_i\).

Subsequently, we project the up-dimensioned feature vectors from \(\mathcal{D}_{\mathrm{train}}\) into the pre-shape space using the procedures described in Eqs.~\eqref{eq:sub_mean}–\eqref{eq:projection}.
Replacing the original features with their pre-shape projections yields
\begin{equation}
    \label{eq:d_ps}
    \mathcal{D}_{\mathrm{ps}} = \left\{\,\bigl(\vect_{i},\,c\bigr) \mid i = 1, \ldots, m \right\}
\end{equation}
where \(\vect_i \in S_*^{2d-3}\), and \( \mathcal{D}_{\mathrm{ps}} \) denotes the set of training features projected into the pre-shape space.

To construct a class-specific Geodesic curve in the pre-shape space, we require two end vectors as learnable parameters for each class. We randomly select two distinct feature vectors from $\left\{ \vecv_{1}, \ldots, \vecv_m \right\}$ to initialize the parameters, denoted as $\vecv_{\mathrm{start}}, \vecv_{\mathrm{end}} \in \mathbb{R}^d$. These vectors are then projected into the pre-shape space, yielding the initial learnable end vectors  $\vect_{\mathrm{start}}, \vect_{\mathrm{end}} \in S_*^{2d-3}$ for subsequent optimization.

\subsection{Geodesic Curve Fitting}
\label{sec:GeodesicCurveFitting}

The FAAGC module aims to obtain the Geodesic curve in the pre-shape space that best fits the given sample set. As illustrated in the \ding{173} of Fig.~\ref{fig:adaptive-geodesic}, to achieve this goal, we optimize the two endpoint vectors of the Geodesic curve, $\vect_{\mathrm{start}}$ and $\vect_{\mathrm{end}}$, as well as the sampling parameter $\boldsymbol{t}$, by designing a loss function and performing gradient descent operations.

During sampling for feature augmentation, we adopt the resampled form of Eq.~\eqref{eq:intermediatepoints}, where the parameter $s$ is normalized to $z = s / \theta$ with $z \in [0,1]$. This normalization restricts the range of the sampling parameter, thereby controlling the magnitude of variation in the subsequent loss calculation. The resampled interpolation formula is given as follows:

\begin{equation}
    f_{\text{interp}}(\vect_{\mathrm{start}}, \vect_{\mathrm{end}}, z) =  \frac{\sin\left[(1 - z) \theta\right]}{\sin \theta} \vect_{\mathrm{start}} + \frac{\sin\left(z \theta\right)}{\sin \theta} \vect_{\mathrm{end}}.
    \label{eq:interp}
\end{equation}

The equivalence between the original and the resampled formulations is proved in Appendix~\ref{sec:equality}.

During optimization, the sampling parameters are treated as a set of learnable parameters of size \(m\), where each element lies in the range \([0, 1]\), denoted as \(\boldsymbol{t} = \{ t_1, t_2, \ldots, t_m \}\). The sampled vectors obtained during training are denoted as \(\matt^{*} = [\vect^{*}_1, \ldots, \vect^{*}_i, \ldots, \vect^{*}_m]\), where \(\vect^{*}_i = f_{\text{interp}}(\vect_{\mathrm{start}}, \vect_{\mathrm{end}}, t_i)\). Here, \(\matt^{*}\) represents a matrix formed by arranging the \(m\) sampled vectors as column vectors.

To ensure that the newly generated sample vectors closely match the distribution of the original vectors \(\matt = [\vect_1, \ldots, \vect_i, \ldots, \vect_m]\), we aim to minimize the Geodesic distances between them. The goal requires adjusting either the sampling parameters \(\boldsymbol{t}\) or the endpoint vectors \(\vect_{\mathrm{start}}\) and \(\vect_{\mathrm{end}}\) of the Geodesic curve to achieve the best possible fit.

A critical issue arises during the Geodesic Curve Fitting stage when treating only $\vect_{\mathrm{start}}$ and $\vect_{\mathrm{end}}$ as optimizable parameters, while fixing $\boldsymbol{t}$ to a set of values pre-sampled from the distribution used for final data augmentation: since the position of the Geodesic curve is not determined until the end of training, it is impossible to align the order of the sample vectors along the final Geodesic curve in advance. During training, the order of the sampling parameters \(\boldsymbol{t} = \{ t_1, t_2, \ldots, t_m \}\) may not correspond to the Geodesic ordering of \(\matt = [\vect_1, \vect_2, \ldots, \vect_m]\).

\begin{figure*}[!t]
    \centering
    \includegraphics[width=0.85\linewidth]{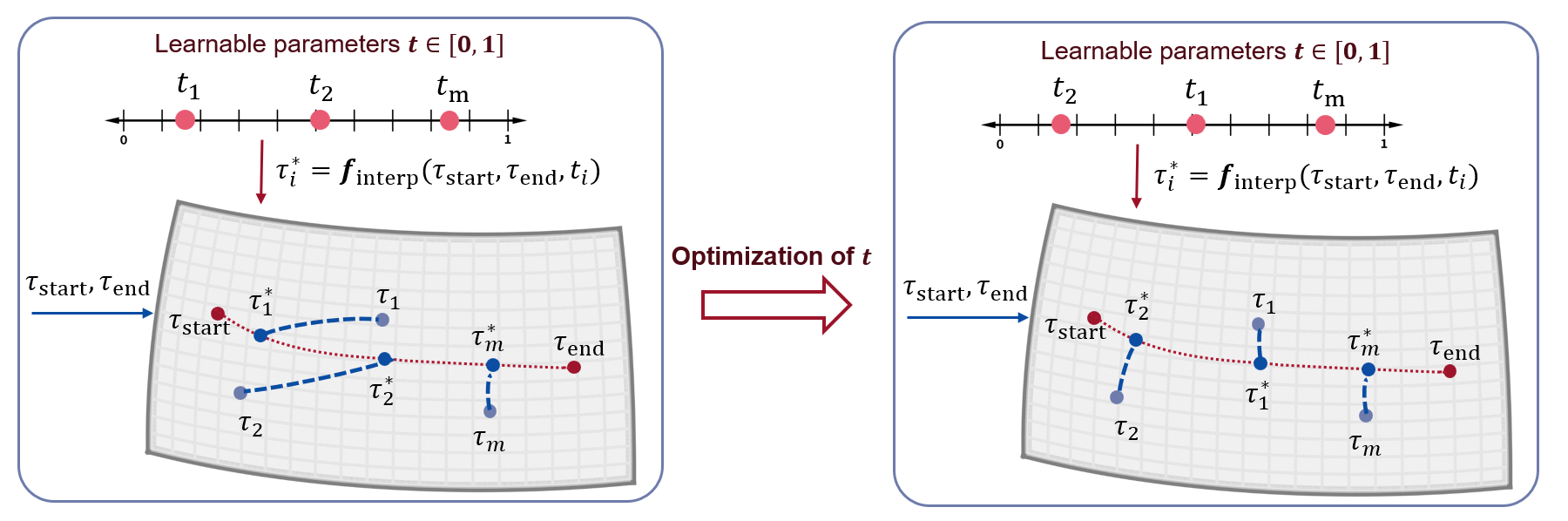}
    \caption{Learnable sampling parameters $\boldsymbol{t}$ allow adaptive alignment between sampled points and original samples, avoiding mismatched pairings and reducing loss.}
    \label{fig:paramt}
\end{figure*}

If a fixed distribution is used during training and the Geodesic distances between the sampled vectors \([\vect^{*}_1, \ldots, \vect^{*}_m]\) and the original vectors \([\vect_1, \ldots, \vect_m]\) are directly minimized, inconsistencies may arise. Even when the Geodesic curve defined by \(\vect_{\mathrm{start}}\) and \(\vect_{\mathrm{end}}\) fits the training vectors well, index mismatches can cause many distances \(d(\vect^{*}_i, \vect_i)\) to remain large. This often leads to unstable training or convergence to suboptimal solutions, as shown in the left part of Fig.~\ref{fig:paramt}.

One approach to address the issue is to avoid computing Geodesic distances based on one-to-one assignments between sampled vectors and original vectors. Instead, all pairwise Geodesic distances between sampled vectors and original vectors can be computed, and their total sum can be minimized as the loss function. However, this computational approach has significant drawbacks: first, it requires a substantial amount of additional computation for Geodesic distances; second, during optimization, each sampled vector \(\{\vect^{*}_i\}\) would tend to move toward regions that minimize Geodesic distances to all original vectors \([\vect_1, \ldots, \vect_m]\), which could lead to excessive concentration of the sampled vectors generated during data augmentation.

By optimizing \(\boldsymbol{t}\), the correspondence between sampled vectors and original vectors can be properly aligned, as illustrated in the right part of Fig.~\ref{fig:paramt}. The sum of loss terms related to the Geodesic distances between each pair of sampled and original vectors is denoted as \(\mathcal{L}_{\text{Sim}}(\matt^{*}, \matt)\).

Although all elements of $\boldsymbol{t}$ are constrained to \([0, 1]\), the underlying distribution from which $\boldsymbol{t}$ is sampled is not directly known. In the final sampling task, it is desirable to freely adjust the scale of the augmented features, which requires sampling parameters \(\boldsymbol{z}\) from a known distribution, such as \(\boldsymbol{z} \sim U(0, 1)\). To ensure that the distribution of the learnable sampling parameters $\boldsymbol{t}$ aligns closely with the distribution of the sampling parameters \(\boldsymbol{z}\) used during data augmentation by the end of training, an additional loss term is designed to minimize the divergence between them. This loss term is denoted as \(\mathcal{L}_{\text{Diverg}}(\boldsymbol{t}, \boldsymbol{z})\).

During training, directly treating \(\vect_{\mathrm{start}}, \vect_{\mathrm{end}} \in S_{*}^{2d-3}\) as model parameters contains risks that these vectors may no longer lie in the pre-shape space due to gradient updates. To prevent this, \(\vecv_{\mathrm{start}}, \vecv_{\mathrm{end}} \in \mathbb{R}^d\), initialized by randomly selecting sample features from the Euclidean space as described in sec~\ref{sec:feat_extraction}, are used as the initial parameters. Their projections into the pre-shape space, denoted as \(\vect_{\mathrm{start}}, \vect_{\mathrm{end}} \in S_{*}^{2d-3}\), are treated as intermediate computational results or activation parameters. This approach ensures that \(\vect_{\mathrm{start}}\) and \(\vect_{\mathrm{end}}\) remain constrained to the pre-shape space throughout gradient-based training.

In summary, the overall loss function consists of two components: the sample similarity loss and the sampling parameter distribution divergence loss. Optimization is performed by minimizing these two loss terms.

The first term is the sample similarity loss $\mathcal{L}_{\text{Sim}}$. By minimizing the Geodesic distance in the pre-shape space between each pair of augmented samples $\vect^{*}_{i}$ (obtained from $\vect_{\mathrm{start}}, \vect_{\mathrm{end}}, \boldsymbol{t}$) and original samples $\vect_i$ in $\matt^{*} = [\vect^{*}_1, \ldots, \vect^{*}_m]$ and $\matt = [\vect_1, \ldots, \vect_m]$, it is ensured that the augmented samples remain close to the original distribution during training. While the Geodesic distance between each $\vect^{*}_{i}$ and $\vect_i$ could be directly used as the loss, cosine similarity is adopted here to simplify computation, as shown in Eq.~\eqref{eq:appro_Sim_loss}:

\begin{equation}
    \label{eq:appro_Sim_loss}
    \mathcal{L}_{\text{Sim}}(\matt^{*},\matt) = \sqrt{ \sum_{i=1}^m \left(1 -  {\vect^{*}_{i}}^\top \vect_i  \right)^2 }.
\end{equation}

The above optimization approach effectively achieves Geodesic Curve Fitting to the training samples, but when the sample size \(m\) and feature dimension \(d\) are large, the computational time may increase significantly. To address this, the calculation process of the loss term is redesigned to enable efficient parallelization.

Specifically, when computing \(\matt^{*} = [\vect^{*}_1, \ldots, \vect^{*}_m]\), Eq.~\eqref{eq:interp} is parallelized and denoted as \(\matt^{*} = f_{\text{interp}}(\vect_{\mathrm{start}}, \vect_{\mathrm{end}}, \boldsymbol{t})\), where the \(i\)-th component \(t_i\) of \(\boldsymbol{t}\) corresponds to the \(i\)-th column vector \(\vect^{*}_i\) of \(\matt^{*}\).

The sample similarity loss \(\mathcal{L}_{\text{Sim}}\) is rewritten in an equivalent matrix operation form as follows:

\begin{equation}
    \mathcal{L}_{\text{Sim}}(\matt^{*},\matt) = \left\| \mathbf{1}_m -  (\matt^{*} \odot \matt)^{\top} \mathbf{1}_{2d}  \right\|,
\end{equation}
where the Hadamard product $\odot$ denotes the element-wise multiplication of two matrices of the same shape, producing a matrix of identical dimensions. Both \(\matt^{*}\) and \(\matt\) are matrices of size \(2d \times m\), where \(m\) denotes the number of sample vectors and \(2d\) represents the dimensionality of each vector. $\mathbf{1}_{2d}$ is a column vector of length $2d$ with all entries equal to one, and $\mathbf{1}_{m}$ is a length-$m$ column vector of ones. The operation \((\matt^{*} \odot \matt)^{\top} \mathbf{1}_{2d}\) (denoted as \((\triangle)\)) corresponds to performing parallel inner products through matrix operations, resulting in a column vector composed of \(m\) inner product values. Subsequently,
\(\left\| \mathbf{1}_m - (\triangle) \right\|\)
represents subtracting these inner products from $1$ and then computing the Euclidean norm, which is equivalent to the process in Eq.~\ref{eq:appro_Sim_loss}, where each of the \(m\) inner product values is subtracted from $1$, squared, summed, and finally square-rooted.

The reformulated loss function enables highly efficient parallel computation in mainstream machine learning libraries such as PyTorch.

The second term is the sampling parameter distribution divergence loss $\mathcal{L}_{\text{Diverg}}$, which measures the discrepancy between the actual distribution $\boldsymbol{z}$ adopted during data augmentation and the learnable sampling distribution $\boldsymbol{t}$. Minimizing $\mathcal{L}_{\text{Diverg}}$ ensures that the learnable parameters $\boldsymbol{t}$ do not deviate excessively from the augmentation distribution $\boldsymbol{z}$ after training, while still allowing greater sampling flexibility during optimization, thereby improving both parameter search efficiency and the final fitting performance.

In implementation, the sampling parameters $\boldsymbol{t}$ are initialized as $\boldsymbol{t}\sim\mathcal{N}(0,I)$ and passed through a sigmoid function to ensure $t_i \in (0,1)$. We adopt $z_i \sim U(0, 1)$ for consistency with prior work~\cite{han2023gcfa}, which also prevents $\boldsymbol{t}$ from collapsing into overly concentrated values that would reduce Geodesic sampling to local point sampling. The Wasserstein distance~\cite{frogner2015learning} is then used to measure the discrepancy between $\boldsymbol{t}$ and $\boldsymbol{z}$, formulated as:

\begin{equation}
    \label{eq:appro_div_loss}
    \mathcal{L}_{\text{Diverg}}(\boldsymbol{t}, \boldsymbol{z}) =  \frac{1}{m} \sum_{j=1}^m \left| t_{(j)} - z_{(j)} \right|
\end{equation}
where $t_{(j)}$ and $z_{(j)}$ denote the $j$-th smallest elements in the sorted sequences of $\boldsymbol{t}$ and $\boldsymbol{z}$, respectively. In Eq.~\eqref{eq:appro_div_loss}, $\mathcal{L}_{\text{Diverg}}(\boldsymbol{t}, \boldsymbol{z})$ attains its minimum value of $0$ only when the sorted sequences of $\boldsymbol{t}$ and $\boldsymbol{z}$ are exactly matched, i.e., $t_{(j)} = z_{(j)}, \quad j=1, \ldots, m$.

For the parallelized implementation of $\mathcal{L}_{\text{Diverg}}(\boldsymbol{t}, \boldsymbol{z})$, once $\boldsymbol{t}$ and $\boldsymbol{z}$ are sorted, the subsequent element-wise absolute differencing and summation can be fully parallelized, thereby greatly improving computational efficiency.

To balance the effects of the two loss terms described above, we introduce a weighting factor $\beta$, and define the overall loss function as follows:

\begin{align}
    \label{eq:train_loss}
    \mathcal{L}_{\text{train}} = \mathcal{L}_{\text{Sim}}(\matt^{*},\matt) + \beta \cdot
    \mathcal{L}_{\text{Diverg}}(\boldsymbol{t}, \boldsymbol{z}).
\end{align}

We employ the Adam optimizer~\cite{kingma2014adam} to optimize $\mathcal{L}_{\text{train}}$. To flexibly control the update rates of different parameters, we assign the learning rate $\eta_p$ to $\vecv_{\mathrm{start}}$ and $\vecv_{\mathrm{end}}$, and the learning rate $\eta_t$ to $\boldsymbol{t}$.

The complete procedure of Geodesic Curve Fitting is outlined in Algorithm~\ref{alg:FAAGC}.

\begin{algorithm}[tb]
    \caption{Geodesic Curve Fitting}
    \label{alg:FAAGC}
    \begin{algorithmic}[1]
        \REQUIRE
        Sample feature vectors $\matt = [\vect_{1}, \ldots, \vect_{m}]$, $\vect_i \in S_{*}^{2d-3}$;\\
        Initialized vectors $\vecv_{\mathrm{start}},\, \vecv_{\mathrm{end}} \in \mathbb{R}^d$;\\
        Loss-weight hyperparameter $\beta$;\\
        Learning rates hyperparameters $\eta_p$, $\eta_t$
        \ENSURE
        Optimized end vectors $\vect_{\mathrm{start}},\, \vect_{\mathrm{end}} \in S_{*}^{2d-3}$
        \WHILE{not converged}
        \STATE Sample $\boldsymbol{t} \sim \mathcal{N}(0,I)$
        \STATE Project $\vecv_{\mathrm{start}}$ and $\vecv_{\mathrm{end}}$ into pre-shape vectors $\vect_{\mathrm{start}}$ and $\vect_{\mathrm{end}}$\\
        \hspace{1.5em}(see Eqs.~\eqref{eq:duplicate}, \eqref{eq:sub_mean}, \eqref{eq:projection})
        \STATE Transform $\boldsymbol{t} \leftarrow \text{Sigmoid}(\boldsymbol{t})$
        \STATE Generate pre-shape sample vectors $\matt^{*}$:\\
        \hspace{1.5em}$\matt^{*} \leftarrow f_{\mathrm{interp}}\!\bigl(\vect_{\mathrm{start}}, \vect_{\mathrm{end}}, \boldsymbol{t}\bigr)$
        \STATE Calculate similarity loss:\\
        \hspace{1.5em}$\mathcal{L}_{\text{Sim}} = \left\| \mathbf{1}_m - (\matt^{*} \odot \matt)^{\top} \mathbf{1}_{2d}\right\|$
        \STATE Sample $\boldsymbol{z} = [z_1, \ldots, z_m]$, $z_i \sim U(0,1)$
        \STATE Calculate divergence loss after sorting elements of $\boldsymbol{t}$ and $\boldsymbol{z}$:\\
        \hspace{1.5em}$\mathcal{L}_{\text{Diverg}} = \frac{1}{m} \sum_{j=1}^m | t_{(j)} - z_{(j)} |$
        \STATE Calculate total loss:\\
        \hspace{1.5em}$\mathcal{L}_{\text{train}} = \mathcal{L}_{\text{Sim}} + \beta \cdot \mathcal{L}_{\text{Diverg}}$
        \STATE Update $\vecv_{\mathrm{start}},\, \vecv_{\mathrm{end}}$ with learning rate $\eta_p$ and $\boldsymbol{t}$ with learning rate $\eta_t$ via Adam optimizer
        \ENDWHILE
        \RETURN Optimized end vectors $\vect_{\mathrm{start}},\, \vect_{\mathrm{end}}$
    \end{algorithmic}
\end{algorithm}

\subsection{Geodesic Data Sampling}

As illustrated in \ding{174} of Fig.~\ref{fig:adaptive-geodesic}, once the optimal $\vect_{\mathrm{start}}$ and $\vect_{\mathrm{end}}$ for the sample featires have been obtained, we sample $n$ parameters $\{z_1, \ldots, z_n\}$ from $U(0,1)$.
By substituting $\{z_1, \ldots, z_n\}$ into Eq.~\eqref{eq:interp}, we sample along the Geodesic curve and obtain the augmented pre-shape vectors $\{\vect^{*}_1, \ldots, \vect^{*}_n\}$.
Unlike the interpolation step in Sec.~\ref{sec:GeodesicCurveFitting}, the augmentation procedure here allows us to specify an arbitrary sampling scale $n$.

For each subset, \(n\) augmented vectors are generated through Algorithm~\ref{alg:FAAGC}, and pseudo-labels are assigned. Thus, the original data is augmented as follows:
\begin{equation}
    \mathcal{D}_{\mathrm{aug}}
    = \left\{ (\vect^{*}_i, c) \mid i = 1, \ldots, n \right\}.
\end{equation}

Both $\mathcal{D}_{\mathrm{ps}}$(See Sec.~\ref{sec:feat_extraction}) and $\mathcal{D}_{\mathrm{aug}}$ are employed to train the downstream classifier.

\section{Experiments}
\label{sec:experiments}

In this section, we validate the effectiveness and applicability of the FAAGC module through a series of experiments.
First, we introduce the datasets, backbone models, and pre-processing steps used in our study.
Second,  a set of comparative experiments are conducted against multiple feature-level augmentation methods to show that the FAAGC module can effectively enhance model performance when training samples are scarce.
Third, we compare FAAGC with the FAGC~\cite{han2023gcfa}, which also constructs Geodesic curve in the pre-shape space, in terms of runtime and augmentation performance to highlight the advantages of our method in both efficiency and effectiveness.
Finally, we perform an ablation study to confirm the necessity of each component of the pipeline and to examine the impact of hyperparameters on the performance of FAAGC.

\subsection{Experimental Setup}
The FAAGC module is evaluated on five visual benchmark datasets:
\begin{itemize}
    \item \textbf{CIFAR-10}~\cite{krizhevsky2009learning}: Contains 50K training and 10K test images (32\(\times\)32 resolution) with 10 classes.
    \item \textbf{CIFAR-100}~\cite{krizhevsky2009learning}: Contains 50K training and 10K test images (32\(\times\)32 resolution) with 100 classes.
    \item \textbf{CUB-200-2011}~\cite{wah2011caltech}: A fine-grained dataset with 200 bird species (11.8K images) with varying image resolutions.
    \item \textbf{Fashion-MNIST}~\cite{xiao2017fashion}:  A 10-class dataset with 60K training and 10K test grayscale images (28\(\times\)28 resolution).
    \item \textbf{Caltech101}~\cite{fei2004learning}: A 101-category object dataset (\(\sim\)9K images, around 300\(\times\)200 resolution) with manual annotations;
\end{itemize}

To simulate data-limited scenarios, we reduce each training set to 5 samples per class, denoted as CIFAR-10@5, CIFAR-100@5, etc., while using the full test set for evaluation.
In particular, for the Caltech101 dataset, we first randomly select 20 samples per class to form the test set, and then randomly sample 5 instances per class from the remaining data as the training set, denoted as Caltech101@5.

For Backbone Models, we use the pre-trained ViT-tiny, which has a feature dimension of $d=192$, to extract features in most of the experiments. Input images are resized to $224\times 224$ with normalization.

Before feature extraction, we first perform fine-tuning for 100 epochs using the Adam optimizer with a batch size of 128. We apply differential learning rates, setting the learning rate to $1\text{e-}3$ for the classification head and $1\text{e-}5$ for the backbone network.

For the Geodesic Curve Fitting step in the FAAGC module, we set the Geodesic loss weight \(\beta=0.3\).
Optimization is performed using the Adam optimizer with 2000 training epochs; the learning rates are set to 0.0003 for \( \eta_p \), and 0.003 for \( \eta_t \).
These rates are determined via grid hyperparameter search on the reduced CIFAR-100 training set to effectively minimize the loss function~\eqref{eq:train_loss} and these hyperparameters are applied consistently across all experiments.

\subsection{Comparative Analysis}
\label{sec:comparative_analysis}

To evaluate the effectiveness of the FAAGC module for feature augmentation, we conduct experiments on five benchmark datasets, namely CIFAR-10@5, CIFAR-100@5, CUB-200@5, Fashion-MNIST@5, and Caltech101@5. For each dataset, we use the pre-trained ViT-Tiny with resizing and normalization. No additional preprocessing is applied.

When augmenting features, although the FAAGC module is capable of generating an arbitrary number of augmented samples \( n \), for a fair comparison with other methods that cannot produce arbitrary amounts of augmented data, we set \( n = m \), where \( m \) is the number of original samples per class. The augmented features are combined with the original features for classifier training, and the loss weighting strategy from FAGC~\cite{han2023gcfa} is used to balance their contributions during optimization. The loss function is given by:

\begin{equation}
    \mathcal{L} =
    p_g \cdot \mathcal{L}_{\mathrm{CE}}(y, \hat{y}) +
    (1 - p_g) \cdot \left[\mathcal{L}_{\mathrm{CE}}(y, \hat{y}) + \lambda \cdot \mathcal{L}_{\mathrm{CE}}(\tilde{y}, \hat{\tilde{y}})\right],
    \label{eq:loss_func}
\end{equation}
where \( p_g \) is set to 0.3 and denotes the probability of training and computing the loss using only \(\mathcal{D}_{\mathrm{ps}}\). The parameter \( \lambda \) is set to 0.5 and serves as a weighting factor applied to the loss computed on augmented data from \(\mathcal{D}_{\mathrm{aug}}\). Here, \( \mathcal{L}_{\mathrm{CE}} \) represents the cross-entropy loss.

We compare FAAGC with a diverse range of existing feature-level augmentation techniques, including FGSM~\cite{goodfellow2014explaining}, Manifold-Mixup~\cite{verma2019manifold}, SFA-S~\cite{li2021simple}, MoEX~\cite{li2021feature}, Feature-level SMOTE~\cite{liu2024feature}, and FAGC~\cite{han2023gcfa}. Unless otherwise specified, all later experiments are conducted 6 times with different random seeds, and we report the mean performance across these runs.

\begin{table*}[!t]
    \centering
    \caption{Classification Accuracies of Different Feature-Augmentation Methods Across the Datasets}
    \label{tab:compare}
    \begin{tabular}{l ccccc}
        \toprule
        Method              & CIFAR-10@5                      & CIFAR-100@5                     & CUB-200@5                       & Fashion-MNIST@5                 & Caltech101@5                    \\
        \midrule
        No Augmentation     & 84.82$\pm$.00                   & 66.41$\pm$.00                   & 71.10$\pm$.00                   & 75.54$\pm$.00                   & 87.28$\pm$.00                   \\
        FGSM                & 84.84$\pm$.03                   & 66.33$\pm$.17                   & 71.10$\pm$.04                   & 75.73$\pm$.00                   & 87.97$\pm$.00                   \\
        Manifold-Mixup      & 83.25$\pm$.23                   & 66.63$\pm$.10                   & 71.13$\pm$.08                   & 75.73$\pm$.00                   & 87.82$\pm$.06                   \\
        SFA-S               & 84.84$\pm$.08                   & 65.26$\pm$.94                   & 71.19$\pm$.04                   & 75.83$\pm$.01                   & 87.08$\pm$.01                   \\
        MoEx                & 84.86$\pm$.04                   & 66.25$\pm$.14                   & 71.25$\pm$.07                   & 75.73$\pm$.00                   & 88.02$\pm$.00                   \\
        Feature-level SMOTE & 84.81$\pm$.03                   & 66.41$\pm$.10                   & 71.16$\pm$.02                   & 75.72$\pm$.00                   & 87.32$\pm$.03                   \\
        FAGC                & 84.97$\pm$.02                   & 66.39$\pm$.03                   & 71.69$\pm$.08                   & 75.71$\pm$.00                   & 88.72$\pm$.03                   \\
        FAAGC               & \textbf{85.05}$\pm$\textbf{.05} & \textbf{67.87}$\pm$\textbf{.04} & \textbf{71.73}$\pm$\textbf{.10} & \textbf{75.96}$\pm$\textbf{.01} & \textbf{88.12}$\pm$\textbf{.01} \\
        \bottomrule
    \end{tabular}
\end{table*}

As shown in Table~\ref{tab:compare}, FAAGC consistently outperforms all baseline methods under data-limited conditions across all benchmark datasets. Notably, it achieves a classification accuracy of 67.87\% on CIFAR-100@5, representing a 1.46 percentage point improvement over the no-augmentation baseline. This performance gain is observed not only over the Euclidean space augmentation methods, but also over the Geodesic-based FAGC method in the pre-shape space. Similar improvements are observed on other datasets' results.

To evaluate the generalizability of FAAGC-based feature augmentation across different classifiers, we first use the pre-trained ViT-tiny model to extract features from the CIFAR-100@5 dataset. FAAGC is then applied to augment these features, and the resulting augmented features, merged with the original ones, are used to train and test three classifiers: $k$-NN, SVM, and MLP. For comparison, we also evaluate several representative feature-level augmentation methods, including SFA-S, MoEx, Feature-level SMOTE, and FAGC, under the same setting.

\begin{table}
    \centering
    \caption{Classification Performance of Augmentation Methods Across Different Classifiers for the CIFAR-100@5 Dataset}
    \label{tab:classfiers}
    \begin{tabular}{lrrr}
        \toprule
        Method              & $k$-NN         & SVM            & MLP            \\
        \midrule
        No Augmentation     & 61.08          & 63.74          & 57.91          \\
        SFA-S               & 61.08          & 41.99          & 58.03          \\
        MoEx                & 61.60          & 64.44          & 58.35          \\
        Feature-level SMOTE & 61.49          & 63.36          & 58.23          \\
        FAGC                & 62.57          & 63.48          & \textbf{66.28} \\
        FAAGC               & \textbf{62.92} & \textbf{65.01} & 66.23          \\
        \bottomrule
    \end{tabular}
\end{table}

As shown in Table~\ref{tab:classfiers}, FAAGC universally improves classifier performance.

To demonstrate the effectiveness of FAAGC-based feature augmentation with different feature extractors, we apply FAAGC to features extracted from the CIFAR-100@5 training set using various backbone networks. Specifically, four backbones are considered: ResNet-50~\cite{he2016deep}, EfficientNet-B4~\cite{tan2019efficientnet}, ViT-tiny~\cite{dosovitskiy2020vit}, and Swin-Transformer-tiny~\cite{liu2021Swin}. Each backbone is first fine-tuned on CIFAR-100@5, using only resizing and normalization for image transformation. The FAAGC training protocol is kept consistent with previous experiments.

To ensure a fair comparison, we evaluate FAAGC alongside several other feature-level augmentation methods, including FGSM, Manifold-Mixup, SFA-S, MoEx, Feature-level SMOTE, and FAGC. All augmentation methods are applied to features extracted by each backbone and are evaluated under identical experimental settings.

\begin{table}
    \centering
    \caption{Classification Accuracies of Augmentation Methods Across Different Backbone Networks for the CIFAR-100@5 Dataset}

    \begin{tabular}{lrrrr}
        \toprule
        Method              & Resnet         & EfficientNet   & ViT            & Swin-Trans.    \\
        \midrule
        No Augmentation     & 40.20          & 39.23          & 66.41          & 73.01          \\
        FGSM                & 41.13          & 39.11          & 66.33          & 73.11          \\
        Manifold-Mixup      & 41.10          & 39.82          & 66.63          & \textbf{73.34} \\
        SFA-S               & 37.40          & 38.99          & 65.26          & 68.01          \\
        MoEx                & 41.03          & 39.70          & 66.25          & 73.03          \\
        Feature-level SMOTE & 41.06          & 39.71          & 66.41          & 72.91          \\
        FAGC                & 41.49          & 40.73          & 66.39          & 73.17          \\
        FAAGC               & \textbf{41.70} & \textbf{40.97} & \textbf{67.87} & 73.26          \\
        \bottomrule
    \end{tabular}
    \label{tab:backbones}
\end{table}

It can be observed from Table~\ref{tab:backbones} that features extracted by various backbone models can be effectively enhanced using the FAAGC method. This enhancement leads to improved classification accuracy with limited sample availability. The lower baseline accuracy of EfficientNet and ResNet backbones compared to ViT and Swin-Transformers is likely due to differences in pretraining: ResNet and EfficientNet use ImageNet-1k, while ViT and Swin-Transformer are pre-trained on the larger ImageNet-21k and ImageNet-22k datasets.

We conduct an experiment to assess FAAGC's data augmentation performance across varying training set sizes \(m\). From the original CIFAR-10, CIFAR-100, CUB-200, and Fashion-MNIST training sets, we extract subsets containing $m \in \{3, 5, 10, 20\}$ samples per class (for Caltech101, $m \in \{3, 5, 10\}$ due to insufficient samples in some classes after reserving 20 test samples).

For each subset size and dataset combination, the feature augmentation and evaluation procedures follow the same steps as described previously.

\begin{table}[htbp]
    \centering
    \caption{Accuracies with FAAGC Under Different Training-Set Sizes}
    \label{tab:num_each_class}
    \resizebox{\columnwidth}{!}{
        \begin{tabular}{lllll}
            \toprule
            \(m\)         & 3                      & 5                      & 10                     & 20                     \\
            \midrule
            CIFAR-10      & 80.31                  & 85.25                  & 88.71                  & 91.58                  \\
            \small +FAAGC & \textbf{81.33} (+1.02) & \textbf{86.04} (+0.79) & \textbf{88.92} (+0.21) & \textbf{91.76} (+0.18) \\
            CIFAR-100     & 57.83                  & 66.41                  & 74.32                  & 78.50                  \\
            \small +FAAGC & \textbf{59.39} (+1.56) & \textbf{67.87} (+1.46) & \textbf{74.67} (+0.35) & \textbf{78.76} (+0.26) \\
            CUB-200       & 62.53                  & 71.11                  & 76.54                  & 79.86                  \\
            \small +FAAGC & \textbf{65.24} (+2.71) & \textbf{71.73} (+0.62) & \textbf{76.71} (+0.17) & \textbf{80.32} (+0.46) \\

            Fashion-MNIST & 71.13                  & 75.54                  & 77.30                  & 81.70                  \\
            \small +FAAGC & \textbf{71.82}(+0.69)  & \textbf{75.96}(+0.42)  & \textbf{77.65}(+0.35)  & \textbf{81.86}(+0.16)  \\
            Caltech101    & 80.93                  & 87.28                  & 90.10                  & --                     \\
            \small +FAAGC & \textbf{82.48}(+1.55)  & \textbf{88.12}(+0.84)  & \textbf{90.74}(+0.64)  & --                     \\
            \bottomrule
        \end{tabular}

    }

\end{table}

The results in Table~\ref{tab:num_each_class} show that the proposed data augmentation method consistently improves classification accuracy across different sample sizes. The performance gain is especially pronounced when the number of samples per class is small. For example, the accuracy on CIFAR-100 increases from 57.83\% to 59.39\% when the number of training samples per class is 3. As the number of training samples per class increases to 20, the improvement becomes marginal, rising from 78.50\% to 78.76\%. These results indicate that FAAGC is particularly effective in scenarios with limited data.

\subsection{Runtime and Accuracy Comparison with FAGC}
When compared with FAGC~\cite{han2023gcfa}, FAAGC and FAGC both project features into the pre-shape space, construct Geodesic curve, and sample along the Geodesic curve for data augmentation. Thus, FAAGC can be regarded as an improvement upon the FAGC method. However, the two methods differ in how the optimal pre-shape space endpoints are obtained. While FAGC obtains them by iteratively computing Geodesic distances, FAAGC optimizes end vectors using a tailored loss function and gradient-descent strategy. This design markedly improves computational efficiency during data augmentation and yields better augmentation performance than FAGC.

The following experiment demonstrates the difference in training time between the two methods. CIFAR-10@5 and CIFAR-10@10 subsets are used to select the optimal end vectors. The original samples are represented by features extracted using the ViT-tiny model. We apply both the FAGC and FAAGC methods for feature augmentation, record the training time for each method, and evaluate the classification accuracy after data augmentation using the same $k$-NN classifier with $k=5$. The feature augmentation steps for both methods is performed on an Intel(R) Xeon(R) Silver 4210R CPU. The experimental results are summarized in Table~\ref{tab:time_complexity}.

\begin{table}[ht]
    \centering
    \caption{Comparison of Training Time and Classification Accuracies between FAGC and FAAGC}
    \label{tab:time_complexity}
    \begin{tabular}{llrr}
        \toprule
        Method & Samples per Class & Training Time (s) & Accuracy (\%) \\
        \midrule
        FAGC   & 5                 & 351.83            & 85.20         \\
        FAAGC  & 5                 & 39.39             & 85.84         \\
        FAGC   & 10                & 711.85            & 86.30         \\
        FAAGC  & 10                & 39.34             & 88.34         \\
        \bottomrule
    \end{tabular}

\end{table}

The results demonstrate that FAAGC achieves higher classification accuracy after feature augmentation, while requiring significantly less computation time. This efficiency is primarily due to the improved Geodesic Curve Fitting process and the efficient loss computation method introduced in Sec.~\ref{sec:GeodesicCurveFitting} . In contrast, FAGC relies on computationally expensive operations, such as extensive point sampling and Geodesic distance calculations, leading to a longer runtime.

\subsection{Compatibility, Ablation, and Hyperparameter Study}

To examine the compatibility of the FAAGC module with various image augmentation techniques during fine-tuning, we conducted experiments on CIFAR-100@5 dataset using following image augmentation strategies:
\begin{enumerate}
    \item No data augmentation is applied; only image resizing and normalization are performed. This setting is used in the previous experiments.
    \item Default ViT augmentation approach, incorporating additional center cropping and random horizontal flipping.
    \item RandAugment~\cite{cubuk2020randaugment} applied during fine-tuning, building upon strategy 2.
    \item AugMix-based~\cite{hendrycks2019augmix} augmentation, also extended from strategy 2.
\end{enumerate}
The features derived from these augmentation methods are either directly classified or augmented via FAAGC before sent to classifier.

\begin{figure}[t]
    \centering
    \includegraphics[width=0.45\textwidth, keepaspectratio]{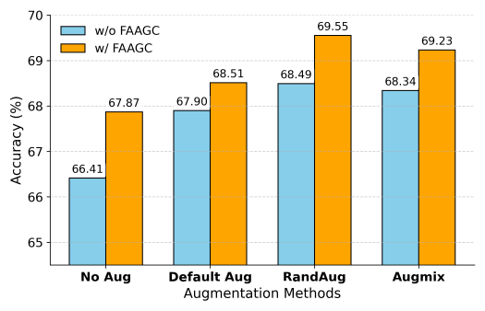}
    \caption{Accuracy Gains from the FAAGC module under different image augmentation methods}
    \label{fig:imageaug_faagc}
\end{figure}

The experimental results from figure~\ref{fig:imageaug_faagc} demonstrate that FAAGC can effectively enhance the performance of deep learning representations under data-limited conditions, regardless of whether image augmentation is applied or not. Moreover, when FAAGC is combined with effective pixel-level augmentation techniques, the classification accuracy can be further improved.

To evaluate the necessity of each stage in the FAAGC pipeline, we conduct ablation experiments on CIFAR-100@5. The ViT-Tiny model is used to extract input features, and we compare three strategies:
\begin{enumerate}
    \item Direct classification on extracted features, training on the \(\mathcal{D_{\mathrm{train}}}\)
    \item Classification after training on the \(\mathcal{D_{\mathrm{ps}}}\)
    \item Classification after applying the complete FAAGC pipeline, training on both \(\mathcal{D_{\mathrm{ps}}}\) and \(\mathcal{D_{\mathrm{aug}}}\).
\end{enumerate}

The hyperparameters utilized for FAAGC remain consistent with those in the previous experiments. We employ ViT-tiny's classification head along with a $k$-NN classifier \((k=5)\) for classification on the test set. The test set classification accuracy results under three different data configurations are presented as table~\ref{tab:aug_ablation} shows:

\begin{table}[t]
    \centering
    \caption{Accuracy Gains From Incremental FAAGC Stages on CIFAR-100@5}
    \label{tab:aug_ablation}
    \begin{tabular}{llrr}
        \toprule
        Projection & Geodesic Augmentation & Head(\%)       & $k$-NN(\%)     \\
        \midrule
        No         & No                    & 67.71          & 61.59          \\
        Yes        & No                    & 68.43          & 61.72          \\
        Yes        & Yes                   & \textbf{69.20} & \textbf{64.69} \\
        \bottomrule
    \end{tabular}

\end{table}

The results in Table~\ref{tab:aug_ablation} demonstrate that the complete FAAGC pipeline achieves the highest accuracy (69.20\% for ViT and 64.69\% for $k$-NN), outperforming both the baseline (67.71\% and 61.59\%) and the partial augmentation variants. Note that the baseline scores in this and subsequent experiments may differ slightly from those reported earlier, as the features are extracted from different fine-tuning runs of the backbone model, even though the same fine-tuning parameters are used in all cases.

To further validate the robustness of FAAGC, we conduct hyperparameter sensitivity studies:
\begin{itemize}
    \item For the weighting parameter \( \beta \) in the loss function~\eqref{eq:train_loss}, we evaluate classification accuracy with
          \[
              \beta \in \{0,\, 10^{-2},\, 3\times 10^{-2},\, 10^{-1},\, 3\times 10^{-1},\, 1,\, 3\}.
          \]
    \item For the learning rate hyperparameters  \( \eta_p \) and  \( \eta_t \), classification accuracy is evaluated with
          \[
              \eta_p  \in \{10^{-4},\, 3\times 10^{-4},\, 10^{-3},\, 3\times 10^{-3},\, 10^{-2},\, 3\times 10^{-2}\},
          \]
          \[
              \eta_t  \in \{10^{-4},\, 3\times 10^{-4},\, 10^{-3},\, 3\times 10^{-3},\, 10^{-2},\, 3\times 10^{-2}\}.
          \]
\end{itemize}

The classification performance is assessed through the ViT classification head and $k$-NN classifier.
\begin{figure}[htb]
    \centering
    \includegraphics[width=0.45\textwidth, keepaspectratio]{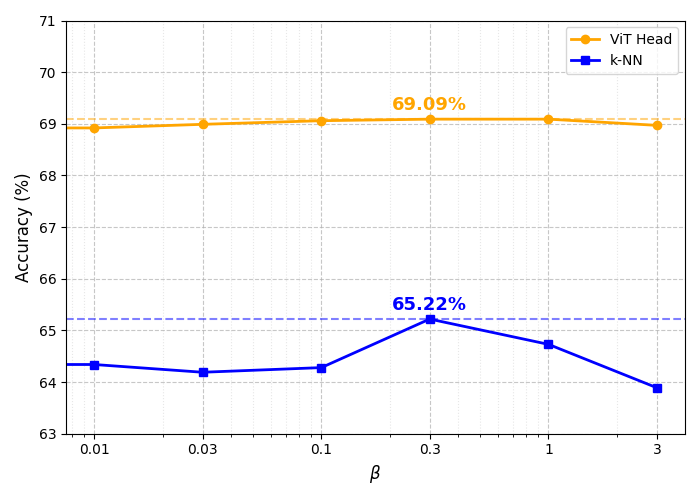}
    \caption{ Effect of the loss-function weight \( \beta \) on classification accuracy.}
    \label{fig:beta_ablation}
\end{figure}

\begin{figure*}[!h]
    \centering
    \includegraphics[width=0.95\linewidth]{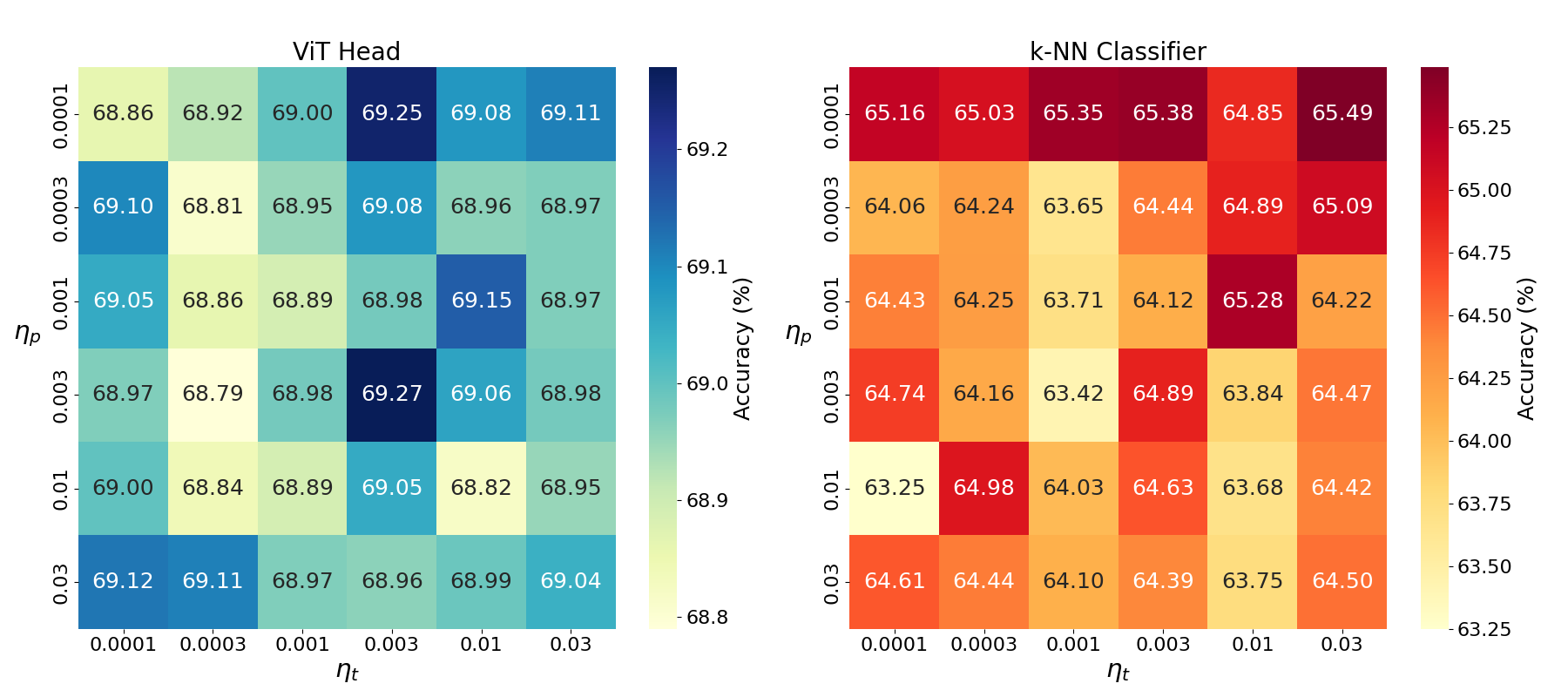}
    \caption{Effect of learning rates \(\eta_p\) and  \(\eta_t\) on classification accuracy}
    \label{fig:lr_ablation}
\end{figure*}

Figure~\ref{fig:beta_ablation} shows that although the magnitude of improvement varies, both classifiers achieve their maximum accuracy gains when $\beta = 0.3$, with the ViT classification head reaching 69.09\% and the $k$-NN classifier attaining 65.22\%. This demonstrates that incorporating the divergence loss with an appropriate weighting coefficient can enhance the sample generation quality of FAAGC.

As evidenced by the heatmap in Figure~\ref{fig:lr_ablation}, the ViT classification head demonstrates robustness to FAAGC, maintaining stable augmentation performance and classification accuracy across varying learning rates. In contrast, the $k$-NN classifier achieves more significant accuracy improvement when utilizing augmented samples generated with lower $\eta_p$ and higher $\eta_t$ values.

Finally, to evaluate the impact of the number of augmented samples per class \(n\) generated by FAAGC on classification performance, we conduct experiments under different augmentation settings. Specifically, we vary the number of synthetic samples per class as \( n \in  \{0, 3, 5, 10, 20, 50 \} \), corresponding to no augmentation and increasing levels of augmentation, respectively. Classification performance is evaluated using both a ViT classification head and a $k$-NN classifier with $k=5$.

\begin{table}[t]
    \centering
    \caption{Effect of the number of augmented samples per class \(n\) generated by FAAGC on classification accuracy.}
    \label{tab:faagc_aug_num}
    \begin{tabular}{lrrrrrrr}
        \toprule
        \(n\)       & 0     & 3     & 5     & 10    & 20    & 50    \\
        \midrule
        Head (\%)   & 67.11 & 67.80 & 67.94 & 67.91 & 67.96 & 68.01 \\
        $k$-NN (\%) & 60.75 & 62.70 & 63.35 & 65.36 & 66.37 & 66.32 \\
        \bottomrule
    \end{tabular}

\end{table}

The results in Table~\ref{tab:faagc_aug_num} show that increasing the number of augmented samples per class improves classification accuracy, especially when the number of synthetic samples exceeds the original. This improvement is particularly evident for the $k$-NN classifier.

\section{Conclusion}

In this paper, we propose a feature-level data augmentation module, FAAGC, by combining shape space theory and deep neural networks to improve model performance. After projecting the extracted feature vectors with deep neural networks into the pre-shape space, the FAAGC module can generate new data by following the Geodesic curve. The FAAGC module is particularly robust in data-scarce scenarios and offers a theoretically grounded framework for feature-level data augmentation.

However, the FAAGC module has some limitations. First, how to effectively combine image-level augmentation with the FAAGC module remains an open question. Exploring whether transformations and augmentations applied at the image input level can help Geodesic curve better capture the distribution of samples is a promising direction for future research. In addition, the performance of the FAAGC module under Geodesic optimization without freezing the backbone has not yet been explored. Nevertheless, fine-tuning the backbone is often necessary for achieving optimal results in certain tasks, such as domain adaptation, fine-grained image classification, and medical image analysis.

Despite these constraints, FAAGC presents a novel module for feature-level data augmentation. Its simplicity and compatibility with standard training pipelines make it especially valuable for applications with limited data.

    {\appendices
        \section*{Proof of Equivalence Between the Geodesic and Interpolation Forms}
        \label{sec:equality}

        This appendix provides a detailed proof that the Geodesic expression
        \eqref{eq:intermediatepoints}
        \begin{multline}
            \Gamma_{(\vect_{\mathrm{start}}, \vect_{\mathrm{end}})}(s)
            = (\cos s)\,\vect_{\mathrm{start}}  \\
            + (\sin s)\frac{\vect_{\mathrm{end}}
                - \vect_{\mathrm{start}}\cos\theta}{\sin\theta},
        \end{multline}
        where $\theta = d(\vect_{\mathrm{start}}, \vect_{\mathrm{end}})$, is equivalent to the interpolation form
        \eqref{eq:interp}
        \begin{multline}
            f_{\text{interp}}(\vect_{\mathrm{start}}, \vect_{\mathrm{end}}, z)
            = \frac{\sin[(1-z)\theta]}{\sin\theta}\vect_{\mathrm{start}}
            + \frac{\sin(z\theta)}{\sin\theta}\vect_{\mathrm{end}},
        \end{multline}
        with $s = z\theta$.

        Starting from the Geodesic formula,
        \begin{multline}
            \Gamma_{(\vect_{\mathrm{start}}, \vect_{\mathrm{end}})}(s)
            = \\
            (\cos s)\vect_{\mathrm{start}}
            + (\sin s)\frac{\vect_{\mathrm{end}} - \vect_{\mathrm{start}}\cos\theta}{\sin\theta},
        \end{multline}

        We can expand the second term as follows:
        \begin{multline}
            \Gamma_{(\vect_{\mathrm{start}}, \vect_{\mathrm{end}})}(s)
            = \\
            (\cos s)\vect_{\mathrm{start}}
            + \frac{\sin s}{\sin\theta}\vect_{\mathrm{end}}
            - \frac{\sin s}{\sin\theta}\cos\theta \,\vect_{\mathrm{start}},
        \end{multline}

        To further simplify the coefficient of $\vect_{\mathrm{start}}$, we recall the trigonometric identity $\sin(\theta-s) = \sin\theta\cos s - \cos\theta\sin s$. This allows us to rewrite
        \begin{equation}
            \frac{\sin\theta\cos s - \cos\theta\sin s}{\sin\theta}
            = \frac{\sin(\theta-s)}{\sin\theta}.
        \end{equation}
        Therefore, the Geodesic expression becomes
        \begin{equation}
            \Gamma_{(\vect_{\mathrm{start}}, \vect_{\mathrm{end}})}(s)
            = \frac{\sin(\theta-s)}{\sin\theta}\vect_{\mathrm{start}}
            + \frac{\sin s}{\sin\theta}\vect_{\mathrm{end}}.
        \end{equation}

        Finally, substituting $s = z\theta$ with $z \in [0,1]$, we obtain
        \begin{multline}
            \Gamma_{(\vect_{\mathrm{start}}, \vect_{\mathrm{end}})}(s) = \Gamma_{(\vect_{\mathrm{start}}, \vect_{\mathrm{end}})}(z\theta) = \\
            \frac{\sin[(1-z)\theta]}{\sin\theta}\vect_{\mathrm{start}} + \frac{\sin(z\theta)}{\sin\theta}\vect_{\mathrm{end}},
        \end{multline}
        which matches exactly the interpolation function $f_{\text{interp}}(\vect_{\mathrm{start}}, \vect_{\mathrm{end}}, z)$.

        Thus, the original Geodesic parameterization and the interpolation expression are mathematically equivalent.

    }

\bibliographystyle{IEEEtran}
\bibliography{ref}

\vfill

\end{document}